\documentclass[conference]{IEEEtran}
\IEEEoverridecommandlockouts
\usepackage{cite}
\usepackage{amsmath,amssymb,amsfonts}
\usepackage{algorithmic}
\usepackage{graphicx}
\usepackage{textcomp}
\usepackage{xcolor}
\usepackage{geometry}
\usepackage{dirtytalk}
\usepackage{booktabs}
\usepackage{array}
\usepackage{float}
\usepackage[english]{babel}
\usepackage{enumitem}
\usepackage[breaklinks]{hyperref}
\usepackage{url}
\usepackage{breakurl}
\usepackage[most]{tcolorbox}

\def\BibTeX{{\rm B\kern-.05em{\sc i\kern-.025em b}\kern-.08em
    T\kern-.1667em\lower.7ex\hbox{E}\kern-.125emX}}

\newtheorem{definition}{Def.}[section]

\expandafter\def\expandafter\UrlBreaks\expandafter{\UrlBreaks
    \do\a\do\b\do\c\do\d\do\e\do\f\do\g\do\h\do\i\do\j%
    \do\k\do\l\do\m\do\n\do\o\do\p\do\q\do\r\do\s\do\t%
    \do\u\do\v\do\w\do\x\do\y\do\z\do\A\do\B\do\C\do\D%
    \do\E\do\F\do\G\do\H\do\I\do\J\do\K\do\L\do\M\do\N%
    \do\O\do\P\do\Q\do\R\do\S\do\T\do\U\do\V\do\W\do\X%
    \do\Y\do\Z\do\/\do-}

\title{Privacy Preservation through Practical Machine Unlearning}

\author{
    \IEEEauthorblockN{Robert Dilworth}
    \IEEEauthorblockA{
        {Department of Computer Science and Engineering} \\
        {Mississippi State University} \\
        {Starkville, Mississippi, USA} \\
        {\href{https://orcid.org/0009-0005-5497-9810}{rkd103@msstate.edu}}
    }
}

\begin{document}

\maketitle

\begin{abstract}
    Machine Learning models thrive on vast datasets, continuously adapting to provide accurate predictions and recommendations. However, in an era dominated by privacy concerns, Machine Unlearning emerges as a transformative approach, enabling the selective removal of data from trained models. This paper examines methods such as Naive Retraining and Exact Unlearning via the SISA framework, evaluating their Computational Costs, Consistency, and feasibility using the \texttt{HSpam14} dataset. We explore the potential of integrating unlearning principles into Positive Unlabeled (PU) Learning to address challenges posed by partially labeled datasets. Our findings highlight the promise of unlearning frameworks like \textit{DaRE} for ensuring privacy compliance while maintaining model performance, albeit with significant computational trade-offs. This study underscores the importance of Machine Unlearning in achieving ethical AI and fostering trust in data-driven systems.
\end{abstract}


\maketitle

\section{Introduction}
\label{sec:Introduction}

    In today’s world, where data is collected incessantly, preserving privacy is increasingly important. Social media platforms, search engines, and other digital services often store vast amounts of user-generated content. This raises concerns about how this data influences Machine Learning (ML) models and what happens when users request its removal.
    
    Machine Learning operates as an additive process—the more data a model ingests, the more robust its predictions become. By contrast, Machine Unlearning is a subtractive process, where specific data points are removed from a model’s training set, ensuring that the updated model no longer retains the removed data’s influence. This approach seeks to maintain a model’s accuracy and utility while adhering to privacy and ethical concerns.
    
    A practical example illustrates this utility: A user posts sensitive information on social media, only to regret it later. Despite deleting the post, its residual influence might persist in recommendation algorithms or other models trained on the platform’s data. Machine Unlearning addresses this issue by expunging the data’s impact entirely.

    \subsection{Motivating Example}
    \label{subsec:Motivating_Example}
    
        The consequences of inadvertent data sharing can be severe, as highlighted by \textit{Mao et al.} \cite{Mao2013}. Twitter users frequently disclose sensitive information such as:
        
        \begin{enumerate}[label=(Conseq. \arabic*), left=2em]
            \item \textbf{Vacation Tweets}: Announcements about vacations inadvertently reveal when homes will be unoccupied, increasing the risk of burglary.
            \item \textbf{Inebriated Tweets}: Drunken posts may divulge private matters, including sexuality, emotional outbursts, confessions, or evidence of illegal activities.
            \item \textbf{Disease Tweets}: Sharing information about health conditions, like HIV or depression, can have lasting personal and professional repercussions.
        \end{enumerate}
        
        These examples underscore the importance of raising awareness about privacy leaks on social media platforms. In scenarios where users seek to erase such posts, Machine Unlearning offers a comprehensive solution that goes beyond mere deletion.

        \subsubsection{Relevance to Machine Learning Models}
        \label{subsubsec:Motivating_Example}
        
            Social media platforms often use user-generated content to train Machine Learning models, such as recommendation systems. Removing a user’s post raises the challenge of ensuring that its influence is also eliminated from these models without degrading their overall performance.

    \subsection{Paper Structure}
    \label{subsec:Paper_Structure}
    
        The remainder of this paper is organized as follows:
        
        \begin{itemize}
            \item \textbf{Sec. \ref{sec:Preliminaries}}: defines key terms and concepts, offering a foundational understanding of Machine Learning, unlearning methods, and relevant datasets.
            \item \textbf{Sec. \ref{sec:Motivations}}: delves into specific scenarios illustrating the importance of Machine Unlearning, with a focus on its impact on privacy and safety.
            \begin{itemize}
                \item \textbf{Sec. \ref{subsec:Research_Objectives}}: outlines the aims of the study, including an evaluation of unlearning methods and their applicability.
                \item \textbf{Sec. \ref{subsec:Hypothesis}}: presents the research hypothesis.
            \end{itemize}
            \item \textbf{Sec. \ref{sec:Literature_Review}}: offers a survey of related works and the theoretical underpinnings of Machine Unlearning approaches.
            \item \textbf{Sec. \ref{sec:Study_Design_and_Methodology}}: details the experimental setup, including the datasets, unlearning methods (Naive Retraining and Exact Unlearning via SISA), and evaluation metrics used to assess Computational Cost and Consistency.
            \item \textbf{Sec. \ref{sec:Experimentation}}: describes the implementation of the proposed study, including the application of Machine Unlearning techniques to the \texttt{HSpam14} dataset, along with a step-by-step analysis of the unlearning processes.
            \item \textbf{Sec. \ref{sec:Results}}: indicates the outcomes of the experiments, providing a comparative analysis of Naive Retraining and Exact Unlearning via SISA. This section highlights the trade-offs between Computational Cost and Consistency, along with insights into their feasibility for real-world applications.
            \item \textbf{Sec. \ref{sec:Review_of_Research}}: discusses the potential integration of Machine Unlearning into Positive Unlabeled Learning frameworks, emphasizing its utility for partially labeled datasets.  
            \item \textbf{Sec. \ref{sec:Conclusion}}: summarizes the findings, highlighting the contributions, challenges, and future directions for Machine Unlearning research.
        \end{itemize}

\section{Preliminaries}
\label{sec:Preliminaries}

    This section introduces key terms and concepts essential to understanding Machine Unlearning.
    
    \subsection{Definitions}
    \label{subsec:Definitions}
    
        \begin{definition}[\textbf{Machine Learning}]
            Machine Learning develops algorithms that enable computers to learn and improve from experience based on data (\textit{Xu et al.} \cite{Xu2024}).
       \end{definition}
    
        \begin{definition}[\textbf{Machine Unlearning}]
           Machine Unlearning involves selectively removing specific training data points’ influence from a pre-trained Machine Learning model (\textit{Jaman et al.} \cite{Jaman2024}). 
           
           Motivated by privacy, security, and legal regulations, Machine Unlearning is essential for data providers and model owners (\textit{Xu et al.} \cite{Xu2023}).
        \end{definition}
    
        \begin{definition}[\textbf{Random Forest}]
            A Random Forest is an ensemble method that averages predictions from multiple decision trees. Its diversity stems from randomness in data sampling and attribute selection (\textit{Brophy et al.} \cite{Brophy2020}).
        \end{definition}
        
        \begin{definition}[\textbf{Naive Retraining}]
           This method involves retraining the model from scratch after removing unwanted data. While effective, it is computationally intensive and reliant on original training data (\textit{Xu et al.} \cite{Xu2024}).
        \end{definition}
        
        \begin{definition}[\textbf{Exact Unlearning via SISA}]
            The Sharding, Isolation, Slicing, and Aggregation (SISA) framework isolates the influence of each data point within specific model shards, enabling efficient retraining of only affected sub-models (\textit{Xu et al.} \cite{Xu2024}).
        \end{definition}
        
        \begin{definition}[\textbf{Consistency}]
            A measure of similarity between predictions from a baseline model and one that has undergone unlearning (\textit{Xu et al.} \cite{Xu2023}).
        \end{definition}
        
        \begin{definition}[\textbf{Computational Cost}]
           Computational Cost captures the time required to delete data points and retrain affected parts of the model. This metric is crucial for evaluating the feasibility of unlearning methods.
        \end{definition}
        
        \begin{definition}[\textbf{DaRE Forest}]
            An implementation of Random Forest that integrates \textit{SISA} principles for efficient unlearning. \textit{DaRE} leverages caching and hierarchical node structures to minimize computational overhead (\textit{Xu et al.} \cite{Xu2024}).
        \end{definition}
        
        \begin{definition}[\textbf{Right to Be Forgotten}]
            Enabling compliance with privacy laws such as \textit{GDPR} and \textit{CCPA}, Machine Unlearning operationalizes the Right to Be Forgotten by removing personal data and its influence on trained models (\textit{Xu et al.} \cite{Xu2024}).
        \end{definition}
        
        \begin{definition}[\textbf{\texttt{HSpam14} Dataset}]
            A dataset of 14 million annotated tweets used for hashtag-oriented spam research, serving as a resource for evaluating unlearning techniques (\textit{Sedhai et al.} \cite{Sedhai2015}).
        \end{definition}

\section{Motivations}
\label{sec:Motivations}

    Social media’s dual role as a platform for connection and a surveillance tool introduces significant risks, particularly for vulnerable populations such as children. For example, child abductors frequently exploit platforms like Twitter to gather personal details, interact with their targets, and orchestrate crimes. These platforms allow malicious actors to monitor online activity, build false trust, and ultimately endanger their victims. In such a scenario, Machine Unlearning could serve as a critical safety measure. Imagine a parent discovering that their child has inadvertently leaked their address in a tweet. Deleting the tweet is only part of the solution; the parents would also need assurance that any Machine Learning models trained on the platform have forgotten this sensitive information. By implementing Machine Unlearning, social media platforms could fulfill such requests, ensuring that the data’s influence is eliminated from the models, thus enhancing safety and preserving privacy \cite{VoiceandView2024, AHA2020, Goggin2023}.
    
    \subsection{Research Objectives}
    \label{subsec:Research_Objectives}
    
        This study aims to evaluate Machine Unlearning methods--Naive Retraining and Exact Unlearning via SISA--by comparing their efficiency and uniformity when applied to the \texttt{HSpam14} dataset. The goal is to determine their feasibility for real-world applications.
    
    \subsection{Study Rationale}
    \label{subsec:Study_Rationale}
    
        Given the growing demand for privacy-preserving technologies, understanding the practical implications of Machine Unlearning is essential. This study will contribute to the field by:
        
        \begin{itemize}
            \item Addressing privacy concerns in data-driven systems.
            \item Demonstrating compliance with legal frameworks like \textit{GDPR} and \textit{CCPA}.
            \item Enhancing the trustworthiness of Machine Learning systems.
        \end{itemize}
    
    \subsection{Hypothesis}
    \label{subsec:Hypothesis}
    
        \begin{enumerate}[label=\textbf{H\arabic*}:, left=2em]
            \item Researchers have applied Machine Unlearning to privacy protection, security, and regulatory compliance challenges.
            \item Exact Unlearning via SISA will outperform Naive Retraining in terms of Computational Cost, while maintaining comparable Consistency.
        \end{enumerate}

\section{Literature Review}
\label{sec:Literature_Review}

    We will now set out to answer the following research question. 
    
    \begin{enumerate}[label=\textbf{RQ\arabic*}:, left=2em]
        \item \textit{What problems have researchers applied Machine Unlearning to?}
    \end{enumerate}
    
    Likewise, we will provide an unfettered description of both the problem formulation and corresponding Machine Unlearning solutions for various works. If a selected paper doesn't lend itself to being summarized in the previously delineated way, then we will provide a commentary on how its content benefits the study. In performing this step, we hope to situate our study within the current body of relevant literature. To this end, we will query the corpora--\textit{IEEE Xplore}, \textit{ACM Digital Library}, and \textit{Google Scholar}--with the string \texttt{\say{Machine Unlearning} AND \say{Application}}.
    
    \subsection{Cao2015}
    \label{subsec:Cao2015}
    
        The researchers chose four real-world, Machine Learning systems--\textit{LensKit} (a media recommendation system), \textit{Zozzle} (a malware detector), \textit{OSNF} (a spam filter), and \textit{PJScan} (a PDF malware detector)--and revised them to incorporate unlearning. They analytically evaluated their algorithmic revision by quantifying the completeness of unlearning and the timeliness of the process. The crux of the study is determining the ease of adapting existing Machine Learning systems to support unlearning. It demonstrates that not every Machine Learning algorithm can be converted to realize unlearning. This is unfortunate as unlearning, or forgetting systems in general, aim to restore privacy, security, and usability \cite{Cao2015}.
    
    \subsection{WangkunXu2024}
    \label{subsec:WangkunXu2024}
    
        This paper introduces a Machine Unlearning algorithm for a load forecasting model to eliminate the influence of data that is adversarial or contains sensitive information of individuals. 
        
        To balance between unlearning completeness and model performance, \textit{PAMU}, a performance-aware Machine Unlearning algorithm, is proposed by evaluating the sensitivity of local model parameter change using influence function and sample re-weighting. To handle the divergence between statistical and task-aware criteria, the authors propose \textit{TAMU}, task-aware Machine Unlearning \cite{WangkunXu2024}.
        
        The key takeaway from this work is the idea of Adversarial Machine Learning, or data poisoning, being a strong motivating factor for Machine Unlearning outside of the obvious compulsion of seeking heightened privacy.
    
    \subsection{Hu2024}
    \label{subsec:Hu2024}
    
        The authors perform an investigation to study to what extent Machine Unlearning methods can partially or fully leak the data of unlearned information. The researchers propose unlearning attacks that, when given access to the original and unlearned models, can reveal the feature and label information of the unlearned data. Their attack extends to both approximate and exact unlearning methods, which directly relates to our intended study. The paper provides sufficient evidence that while Machine Unlearning is invaluable to preserving privacy, it too can be leveraged to thwart privacy. This is especially relevant if uncovered features, as unearthed in the attack demonstrated in the paper, contain personally identifiable information \cite{Hu2024}.
    
    \subsection{Zhang2024}
    \label{subsec:Zhang2024}
    
        The researchers propose a framework for protecting the deletion rights of dataset owners in Machine Unlearning scenarios. \textit{DuplexGuard} makes use of duplex watermarks, or a set of two watermarks, that allow the framework to signify potential dataset usage statuses. Watermarking, or backdoor watermarking, is commonly used for dataset inference to prove its legitimate ownership by means of making the model memorize specific information other than that from the normal samples, and then use this knowledge for verification afterward \cite{Zhang2024}.
        
        The most compelling contribution of this work is \textit{DuplexGuard}'s ability to verify the unlearning effect of the target model after a data deletion request is issued. It is one thing to claim that a Machine Unlearning implementation deletes elements as requested, but it is far more important to ensure the effect of the removal cascades throughout the affected model. It could be argued that the verification aspect of Machine Learning is inseparable from the approach, for what use does a deletion request serve if it does not achieve some modicum of privacy?
    
    \subsection{Lin2023}
    \label{subsec:Lin2023}
    
        The researchers construct an experiment that makes use of \textit{DaRE}, which we have indicated as a suitable Machine Unlearning implementation for our own study. The authors provide insights into its capabilities, which, again, proves beneficial for our current endeavors. The paper describes \textit{DaRE} like so. \textit{DaRE} only supports binary classification tasks and uses a simple model that considers all trees independently.
        
        The authors juxtapose \textit{DaRE}'s failings with their improved Machine Unlearning implementation: Gradient-Boosting Decision Trees (\textit{GBDT}) with Machine Unlearning functionality. Their \textit{GBDT} model interweaves dependencies within the trees, which, from their results, can lead to superior unlearning performance \cite{Lin2023}.
    
    \subsection{Tarun2023}
    \label{subsec:Tarun2023}
    
        The text delves into the \textit{UNSIR} method, or unlearning by selective impair and repair, that thrives in a zero-glance privacy setting. A zero glance setting is one where data samples of the requested unlearning class is either not available or can not be used, which the authors posit has practical real-world applications. The problem that they address is the use-case of unlearning as it pertains to facial recognition. The results seem to support that the \textit{UNSIR} method can make a trained model efficiently forget a single face or multiple faces without glancing at the samples of the unlearning faces \cite{Tarun2023}.
        
        This strict, self-imposed problem formulation confers greater privacy guarantees, as it assumes that the model owner no longer has access to the material requested for removal. The study not only operates under the assumption of absolute inaccessibility but also ensures that their implementation executes as if it is an iron-clad fact. This is especially desirable given the potential fallout of highly sensitive information.
    
    \subsection{Schelter2021}
    \label{subsec:Schelter2021}
    
        The study examines the problem of low-latency Machine Unlearning, which is concerned with maintaining a deployed ML model in place under the removal of a small fraction of training samples without retraining. The researchers proposed a classification model called \textit{HedgeCut} for this setting, which is based on an ensemble of randomized decision trees \cite{Schelter2021}.
        
        \textit{HedgeCut}, much like \textit{DaRE}, was one of the few Machine Unlearning implementations with open-source code that we found during our scouring of literature. Given how both deal with ensemble methods, with the former working with Extremely Randomised Trees (\textit{ERTs}) and the latter with standard trees. Our decision to prioritize the use of \textit{DaRE} over \textit{HedgeCut} came down to a simple cost-benefit analysis. The learning curve of picking up \textit{DaRE} over \textit{HedgeCut} was smaller, ergo we went with \textit{DaRE}. Despite our determination, we would be remiss not to mention other Machine Unlearning alternatives.
    
    \subsection{Juliussen2023}
    \label{subsec:Juliussen2023}
    
        To visually represent our study's usage of Naive Retraining and Exact Unlearning via Sharding, Isolation, Slicing, and Aggregation (SISA), we provide suitable figures that encapsulate each process. Note that the author's usage of \say{Exact Unlearning} corresponds to our definition of \say{Naive Retraining}, whereas \say{Sharded, Isolated, Sliced, and Aggregated Training} align with our characterization of \say{Exact Unlearning via Sharding, Isolation, Slicing, and Aggregation (SISA)}. In any case, we attribute our usage of the two figures to their rightful owners; they elegantly capture our topic of discussion in easily interpretable diagrams.
        
        \begin{figure}[H]
            \centering
            \includegraphics[width=1\linewidth]{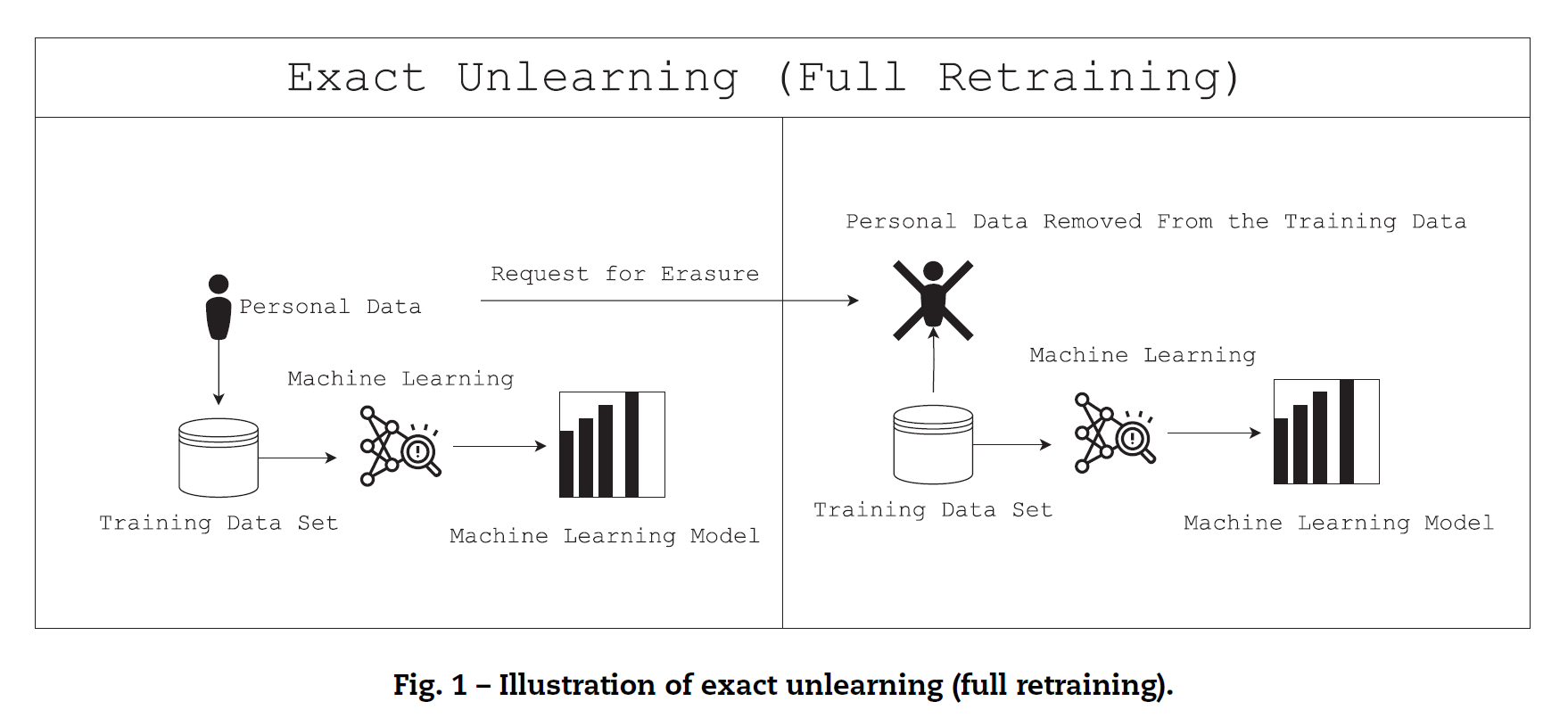}
            \caption{\cite{Juliussen2023}'s Depiction of Naive Retraining}
        \end{figure}
        
        \begin{figure}[H]
            \centering
            \includegraphics[width=1\linewidth]{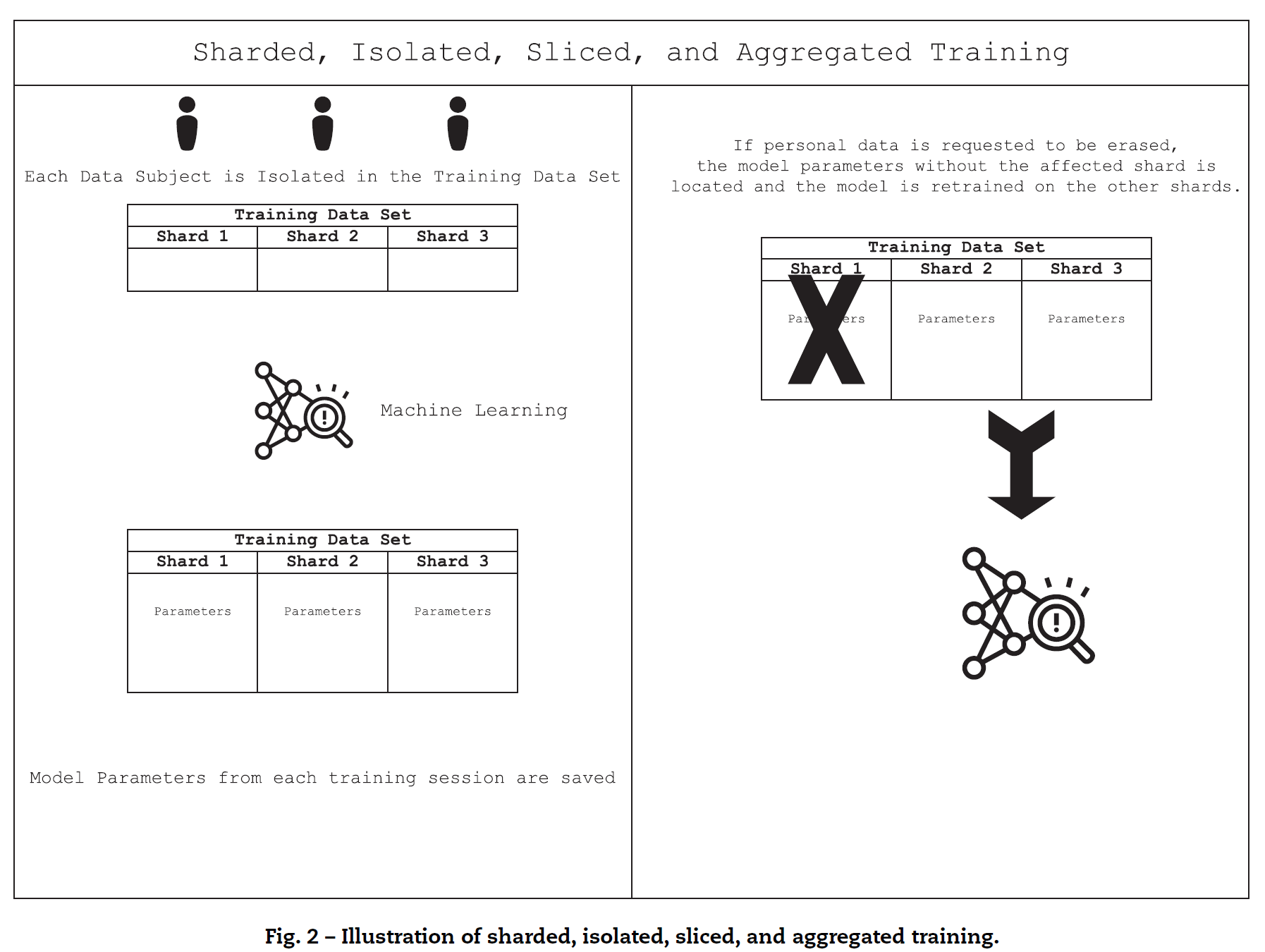}
            \caption{\cite{Juliussen2023}'s Depiction of Exact Unlearning via SISA}
        \end{figure}

    \subsection{Chen2025}
    \label{subsec:Chen2025}

        While Machine Unlearning holds promise as a technique to enhance privacy and data control, the authors identify several potential avenues for misuse by malicious actors \cite{Chen2025}. These include:
        
        \begin{enumerate}[label=(Atk. \arabic*):, left=2em]
            \item \textbf{Misclassification Attacks}: Adversaries manipulate unlearning to cause misclassifications of target samples.
            \item \textbf{Adversarial Robustness Attacks}: Adversaries exploit unlearning to weaken model robustness against adversarial examples.
            \item \textbf{Backdoor Injection Attacks}: Adversaries inject \textit{backdoors} (a poisoned, adversarially crafted data point that alters a model's predictions when a specific trigger is present in its input data) in unlearned models by submitting malicious unlearning requests.
            \item \textbf{Fairness Attacks}: Adversaries craft unlearning sets to introduce bias and discrimination in unlearned models.
            \item \textbf{Explanation Attacks}: Adversaries exploit unlearning to invalidate previously generated model explanations.
            \item \textbf{Running Time Attacks}: Adversaries attack unlearning algorithms to increase the cost of erasing certain requests.
        \end{enumerate}
        
        While the promise of Machine Unlearning is alluring, it would be irresponsible to present an overly \say{rosy} depiction without acknowledging its potential pitfalls. Like any emerging technology, Machine Unlearning carries both merits and demerits that must be carefully weighed.

\section{Study Design and Methodology}
\label{sec:Study_Design_and_Methodology}

    We will be conducting an exploratory study that ascertains the performance of the following Machine Unlearning approaches: Naive Retraining and Exact Unlearning via SISA.
    
    \subsection{Gauging the Computational Cost and Consistency of Machine Unlearning: Naive Retraining versus Exact Unlearning via Sharding, Isolation, Slicing, and Aggregation (SISA)}
    \label{subsec:Gauging}
    
        Our evaluation metrics will be:
        
        \begin{enumerate}[label=(Met. \arabic*):, left=2em]
            \item Consistency (or prediction accuracy) and
            \item Computational Cost (or an aggregate of unlearning and retraining time).
        \end{enumerate}
        
        Given the availability of a publicly available Machine Unlearning implementation, \textit{DaRE}, the experiment will be conducted using a Random Forest Machine Unlearning-enabled model. Likewise, as a consequence of using pre-built code, our programming language of choice will be \textit{Python}. The code that we will utilize was sourced from a repository associated with \textit{Brophy et al.}'s publication \cite{Brophy2020}, which can be found here \cite{Brophy2021}.
        
        \subsubsection{Exact Unlearning via SISA}
        \label{subsubsec:Exact_Unlearning}
        
            First, we will capture the resultant output of a function call of the form, \texttt{rf.predict\_proba(X\_test)}, in addition to the time it took to execute some arbitrary deletion, \texttt{rf.delete(index\_value)}. As a reminder, the function calls will be accessing a special Random Forest model instance, \textit{DaRE}, catered for Machine Unlearning. The first function call provides a prediction; the second removes a set of elements from the training set used to build the Random Forest model. 
            
            In total, both function calls will be timed; the value pair to-be-deleted will be a permissible pseudo-random value that corresponds to a valid index of the dataset under scrutiny (or the training set).
            
            We will use the Twitter dataset, \texttt{HSpam14}, due to its relevance in evaluating privacy concerns associated with user-generated social media data. This dataset serves as a suitable benchmark because it contains annotated tweets, which provide a realistic scenario for evaluating the impact of oversharing on privacy. Social media platforms often exploit such data for training Machine Learning models, raising concerns about the retention of sensitive information. For example, oversharing could lead to data breaches, identity theft, or personal harm, as discussed in Sec. \ref{sec:Motivations}. Therefore, this dataset aligns well with the study's goal of demonstrating the effectiveness of unlearning methods in preserving user privacy.
        
        \subsubsection{Naive Retraining}
        \label{subsubsec:Naive_Retraining}
        
            Next, we will juxtapose the exact Machine Unlearning approach with a naive implementation that does the following. First, it trains a normal Random Forest model. Second, it deletes a random element within the dataset's training set, accounting for both the \textit{x} and \textit{y} components. Finally, it retrains the dataset to produce another Random Forest classifier and outputs a prediction probability using the test set. The timing for this iteration will begin at the second step and terminate at the final step.
            
            Note: Both the Naive Retraining and Exact Unlearning via SISA trials will make use of the same test set for continuity. Likewise, the shared test set will be kept at a relatively constant size throughout the examinations, of which there will be nine. This number stems from the training set configurations that we plan to observe. We plan to gauge the influence of a training set's size and the deletion percentage enacted upon it through the lens of a model's Consistency and Computational Cost. The training set sizes that we will work with are as follows: \textit{n = 10, 100, 1000}. The training set deletion percentages that we will evaluate are \textit{\% = 0.25, 0.50, 0.75}. 
            
            For the first trial, we will retrieve 10 elements from the \texttt{HSpam14} dataset, which will comprise our training set. With these 10 elements, we will then delete a quarter of the elements and capture our evaluation metrics. Next, we will revert the training set back to its original size, delete half the elements, and record our metrics. Finally, we will restore the training set back to its original dimensions, delete three-fourths of the elements, and re-record our metrics. As you can see, for each training set size, \textit{n}, we will apply the three deletion percentages, \textit{\%}, leading to a total of nine trials. Seeing as how we are evaluating two Machine Unlearning approaches, there will be a grand total of 18 trials: 9 for Naive Retraining and 9 for Exact Unlearning via SISA.

\section{Experimentation}
\label{sec:Experimentation}

    In lieu of a more rigid experimental exegesis, we will indicate the executional sequence of our codebase, or the code we crafted to realize the experiment. In this way, we will first begin by dictating the order in which our scripts were ran. After each enumerated script name, we will provide a summary of what the script does, what it accomplishes, and how it goes about realizing our experiment. 
    
    \subsection{Script Execution Pipeline}
    \label{subsec:Script_Execution_Pipeline}
    
        \begin{enumerate}[label=(Scr. \arabic*), left=0.1em]
            \item \texttt{generate\_dataset.py}
            \begin{itemize}
                \item Script Programming Language: 
                \begin{itemize}
                    \item Python
                \end{itemize}
                \item Script Summary: 
                \begin{itemize}
                    \item \texttt{generate\_dataset.py} script is designed to preprocess and consolidate two datasets: a CSV file containing tweets and a text file labeled as \texttt{HSpam14}. The process involves loading the tweet data, extracting and renaming the relevant text column, and saving it in a modified format. It also reads the \texttt{HSpam14} dataset, assigns appropriate column names, and cleans up unnecessary columns. After tidying both datasets, the script concatenates them based on their lengths to ensure compatibility, resulting in a final DataFrame that includes \texttt{user\_id}, \texttt{label}, and \texttt{text}. The consolidated dataset is then saved as a CSV file for future analysis or modeling tasks.
                \end{itemize}
            \end{itemize}
            \item \texttt{preprocess\_dataset.py}
            \begin{itemize}
                \item Script Programming Language:
                \begin{itemize}
                    \item Python
                \end{itemize}    
                \item Script Summary:
                \begin{itemize}
                    \item The \texttt{preprocess\_dataset.py} script is designed to preprocess a dataset of Twitter data for further analysis or Machine Learning tasks. It includes functionalities for feature engineering, data cleaning, and encoding categorical variables. The \texttt{dataset\_specific()} function of \texttt{preprocess\_dataset.py} retrieves and preprocesses the dataset, performing feature engineering and splitting the data into training and testing sets. The main function of \texttt{preprocess\_dataset.py} facilitates the preprocessing of the dataset, including feature encoding and output binarization (the process of transform continuous variables into binary ones). It also handles the saving of the processed data into a specified directory. 
                \end{itemize}
            \end{itemize}
            \item \texttt{evaluate\_metrics.py}
            \begin{itemize}
                \item Script Programming Language:
                \begin{itemize}
                    \item Python
                \end{itemize}
                \item Script Summary:
                \begin{itemize}
                    \item \texttt{evaluate\_metrics.py} implements a comparative analysis of two Machine Unlearning approaches in the context of data deletion: Naive Retraining using a Random Forest Classifier and Exact Unlearning via SISA using a Removal-Enabled Random Forest (or \textit{DaRE-RF}). The primary objective is to evaluate the impact of data removal on model performance, specifically focusing on Consistency (classifier prediction accuracy) and Computational Cost (elapsed time of unlearning).
                    \item Results, including Consistency before and after deletions, Computational Costs, and percentage changes in Consistency, are logged into a text file for further analysis. For clarity, we provide the computation used to obtain the percentage changes in Consistency.
                        \begin{figure}[H]
                        \begin{equation}
                        C_{\Delta} = \left(\frac{C_{after} - C_{before}}{C_{before}}\right) * 100
                        \end{equation}
                        \caption{Percent Change in Consistency Formula: Let \(C_{before}\) be the Consistency reading before unlearning, \(C_{after}\) be the Consistency reading after unlearning, and \(C_{\Delta}\) be the Percent Change in Consistency.}
                        \end{figure}
                    \item The code is designed to run multiple scenarios with varying deletion percentages and target sizes, facilitating a comprehensive evaluation of the two methodologies under different conditions.
                    \item Overall, this code serves as a framework for investigating the effects of data deletion on Machine Learning model performance, contributing to the understanding of model robustness and the implications of data management practices in Machine Learning. 
                \end{itemize}
                \item Function Summaries: 
                \begin{enumerate}[label=(Fcn. \arabic*), left=0.1em]
                    \item \texttt{get\_data()}: This function loads training and testing datasets from .npy files. It assumes that the last column of the training data contains the labels. It returns the feature matrices and label vectors for both training and testing datasets.  
                    
                    \item \texttt{delete\_percentage(X\_train, X\_test, y\_train, y\_test, delete\_percentage)}: This function calculates the number of samples to delete from the training and testing datasets based on the specified \texttt{delete\_percentage}. It randomly selects indices to delete and returns the reduced datasets.  
                    
                    \item \texttt{delete\_n\_elements(X\_train, y\_train, n)}: This function deletes \texttt{n} random elements from the training feature matrix and label vector. It checks if \texttt{n} exceeds the length of the arrays and iteratively removes elements by randomly selecting indices.  
                    
                    \item \texttt{reduce\_to\_target\_size(X\_train, X\_test, y\_train, y\_test, target\_size)}: This function reduces the training and testing datasets to a specified \texttt{target\_size}. It ensures that the size does not exceed the available number of samples and returns the reduced datasets.  
                \end{enumerate}
                \item Steps Executed in the Driver Code, \texttt{main()}:
                \begin{enumerate}[label=(Stp. \arabic*), left=0.1em]
                    \item Print the examination parameters (deletion percentage and target size).  
                    \item Initialize a Stopwatch instance.  
                    \item Load the training and testing datasets using \texttt{get\_data()}.  
                    \item Reduce the datasets to the specified target size using \texttt{reduce\_to\_target\_size()}.  
                    \item Naive Retraining:  
                    \begin{itemize}
                        \item Train a Random Forest Classifier on the reduced training dataset.  
                        \item Predict and calculate Consistency (accuracy) on the testing dataset before deletion.  
                        \item Start the stopwatch.  
                        \item Delete a specified percentage of elements from the training dataset using \texttt{delete\_n\_elements()}.  
                        \item Retrain the Random Forest Classifier on the modified training dataset.  
                        \item Predict and calculate Consistency (accuracy) on the testing dataset after deletion.  
                        \item Stop the stopwatch and calculate Computational Cost (the elapsed time).  
                        \item Log the results (Consistency before and after deletion, Computational Cost) to a text file.
                    \end{itemize}
                    \item Exact Unlearning via SISA:
                    \begin{itemize}
                        \item Train a SISA model using the \texttt{dare} library on the same reduced training dataset.  
                        \item Predict and calculate Consistency (accuracy) on the testing dataset before deletion.  
                        \item Start the stopwatch.  
                        \item Delete a specified percentage of elements from the SISA model. The deletion of a specified percentage of elements from the SISA model is accomplished iteratively through a loop that executes the delete method of the \texttt{dare\_rf} object, using a terminating range value of \texttt{int(target\_size * delete\_percentage\_value)}, which calculates the total number of elements to delete based on the specified \texttt{delete\_percentage\_value} and the \texttt{target\_size}. This ensures that the number of deletions corresponds to the desired percentage of the dataset.  
                        \item Predict and calculate Consistency (accuracy) on the testing dataset after deletion.  
                        \item Stop the stopwatch and calculate Computational Cost (the elapsed time).  
                        \item Log the results (Consistency before and after deletion, Computational Cost) to the same text file.  
                    \end{itemize}
                    \item The main function is called multiple times with different combinations of \texttt{delete\_percentage\_value} and \texttt{target\_size} to perform the analysis under various conditions. Specifically, the target size values used are \textit{n = 10, 100, 1000}, and the deletion percentages are \textit{\% = 0.25, 0.50, 0.75}. This results in a comprehensive evaluation of the model's Consistency and Computational Cost across a range of scenarios, allowing for a detailed understanding of how varying the size of the dataset and the extent of data deletion impacts the effectiveness of both the Naive Retraining and Exact Unlearning via SISA approaches.
                \end{enumerate}
            \end{itemize}
                \begin{figure}[H]
                    \centering
                    \includegraphics[width=1\linewidth]{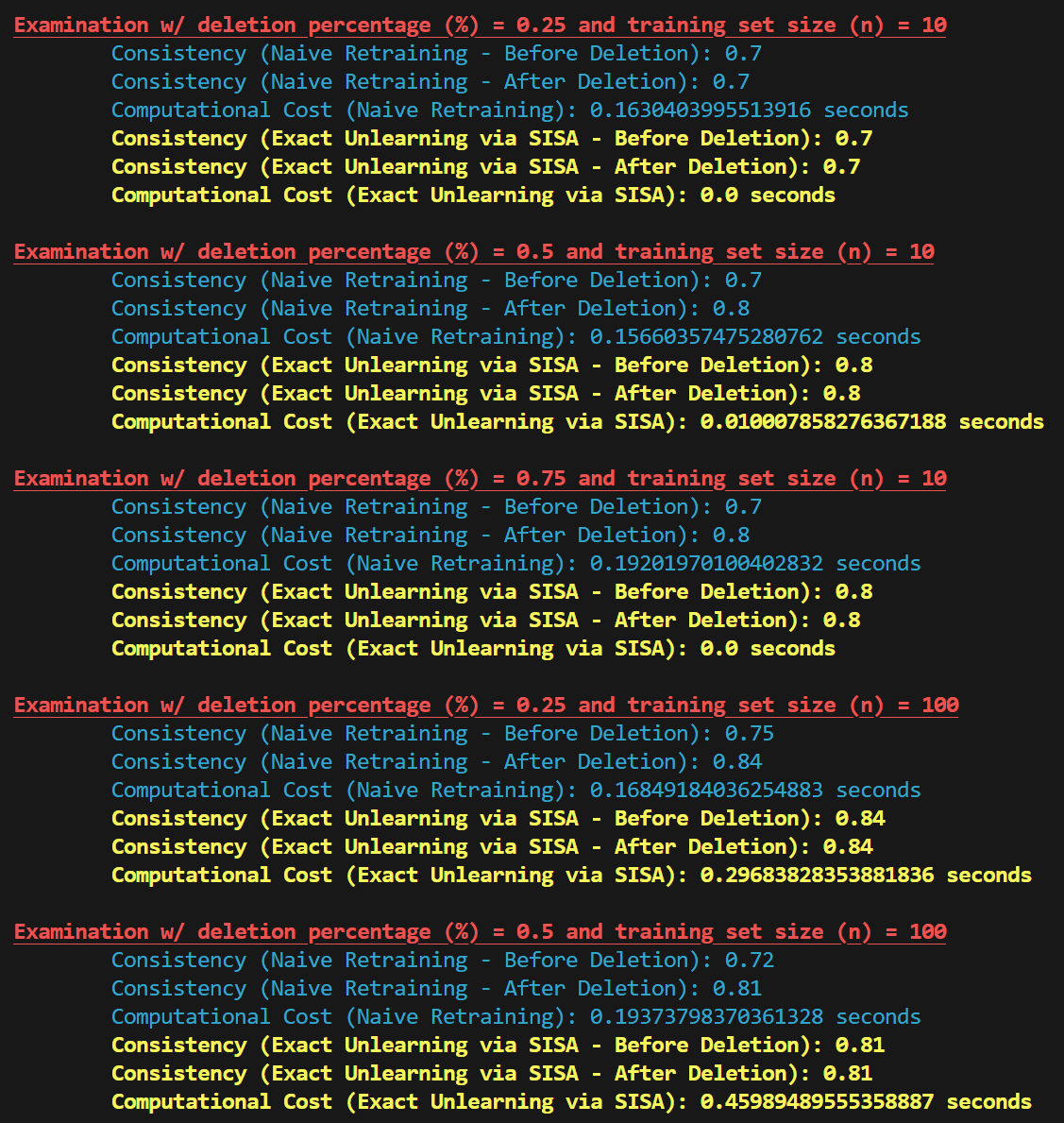}
                    \caption{Script Output 1: \texttt{evaluate\_metrics.py}}
                \end{figure}
                
                \begin{figure}[H]
                    \centering
                    \includegraphics[width=1\linewidth]{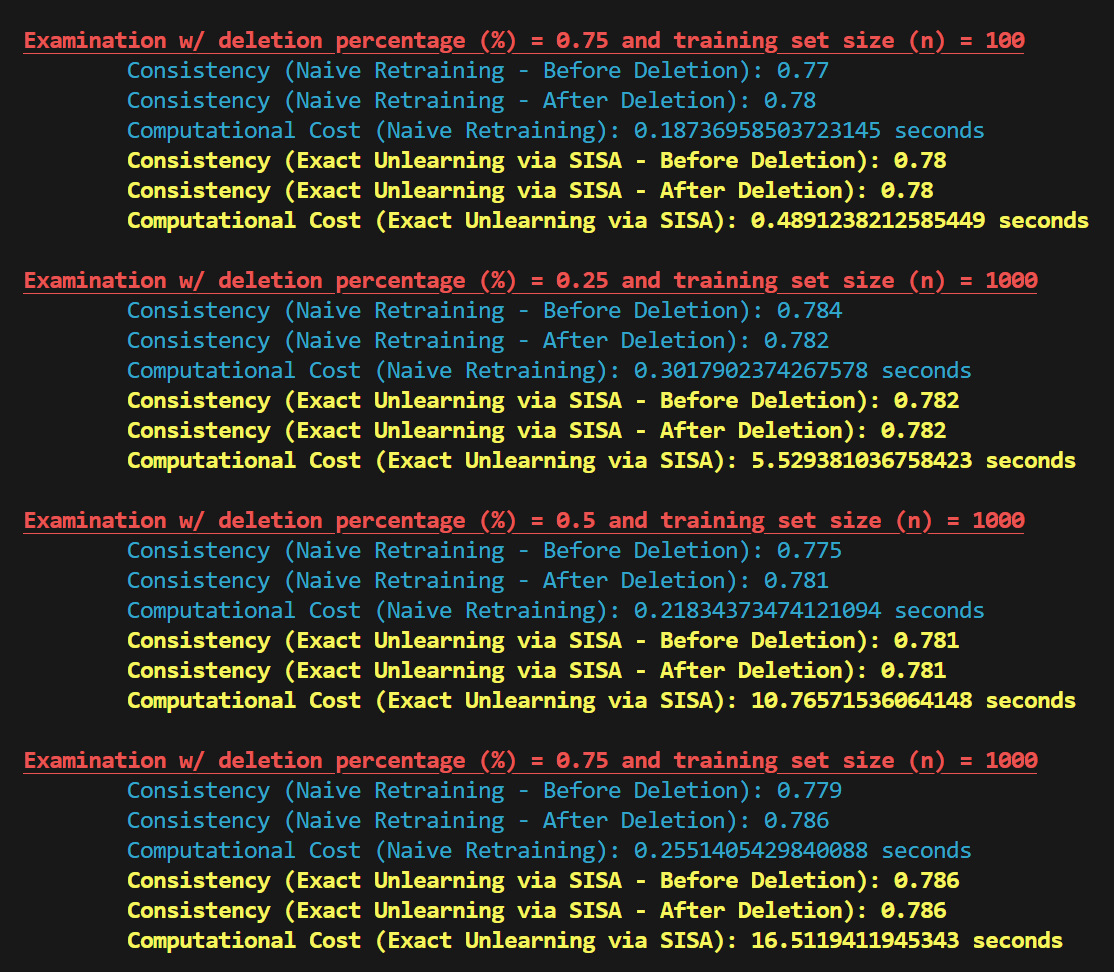}
                    \caption{Script Output 2: \texttt{evaluate\_metrics.py}}
                \end{figure}
            \item \texttt{tidy\_results.py}
            \begin{itemize}
                \item Script Programming Language:
                \begin{itemize}
                    \item Python
                \end{itemize}
                \item Script Summary:
                \begin{itemize}
                    \item \texttt{tidy\_results.py} implements a straightforward data transformation process that converts a text file containing experimental results into a structured CSV format suitable for further analysis.
                \end{itemize}
            \end{itemize}
            \item \texttt{visualize\_results.r}
            \begin{itemize}
                \item Script Programming Language:
                \begin{itemize}
                    \item R
                \end{itemize}
                \item Script Summary:
                \begin{itemize}
                    \item \texttt{visualize\_results.r} implements a data visualization workflow in R, utilizing the \texttt{ggplot2} library to analyze and present the results of a Machine Unlearning experiment. \texttt{visualize\_results.r} produces a visual representation of the Computational Cost and Consistency changes associated with various Machine Unlearning approaches preserved for further analysis and dissemination. This visualization serves as a critical tool for understanding the trade-offs between Computational Cost and Consistency in Machine Unlearning methodologies, specifically comparing the Exact Unlearning via SISA approach against Naive Retraining. For reporting purposes, it also generates visual representations of the original dataset and a glimpse of that dataset.
                \end{itemize}
            \end{itemize}
                \begin{figure}[H]
                    \centering
                    \includegraphics[width=1\linewidth]{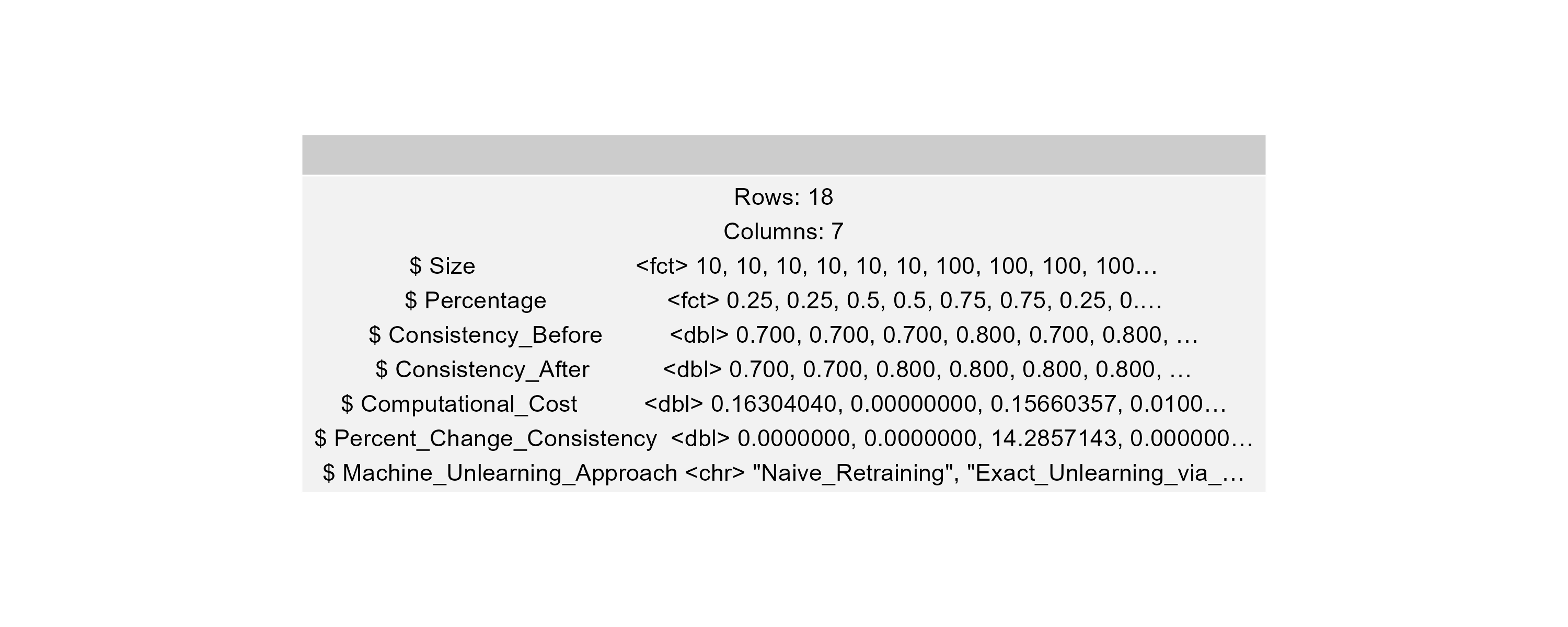}
                    \caption{A Glimpse into our Experimental Result Dataset}
                \end{figure}
                
                \begin{figure}[H]
                    \centering
                    \includegraphics[width=1\linewidth]{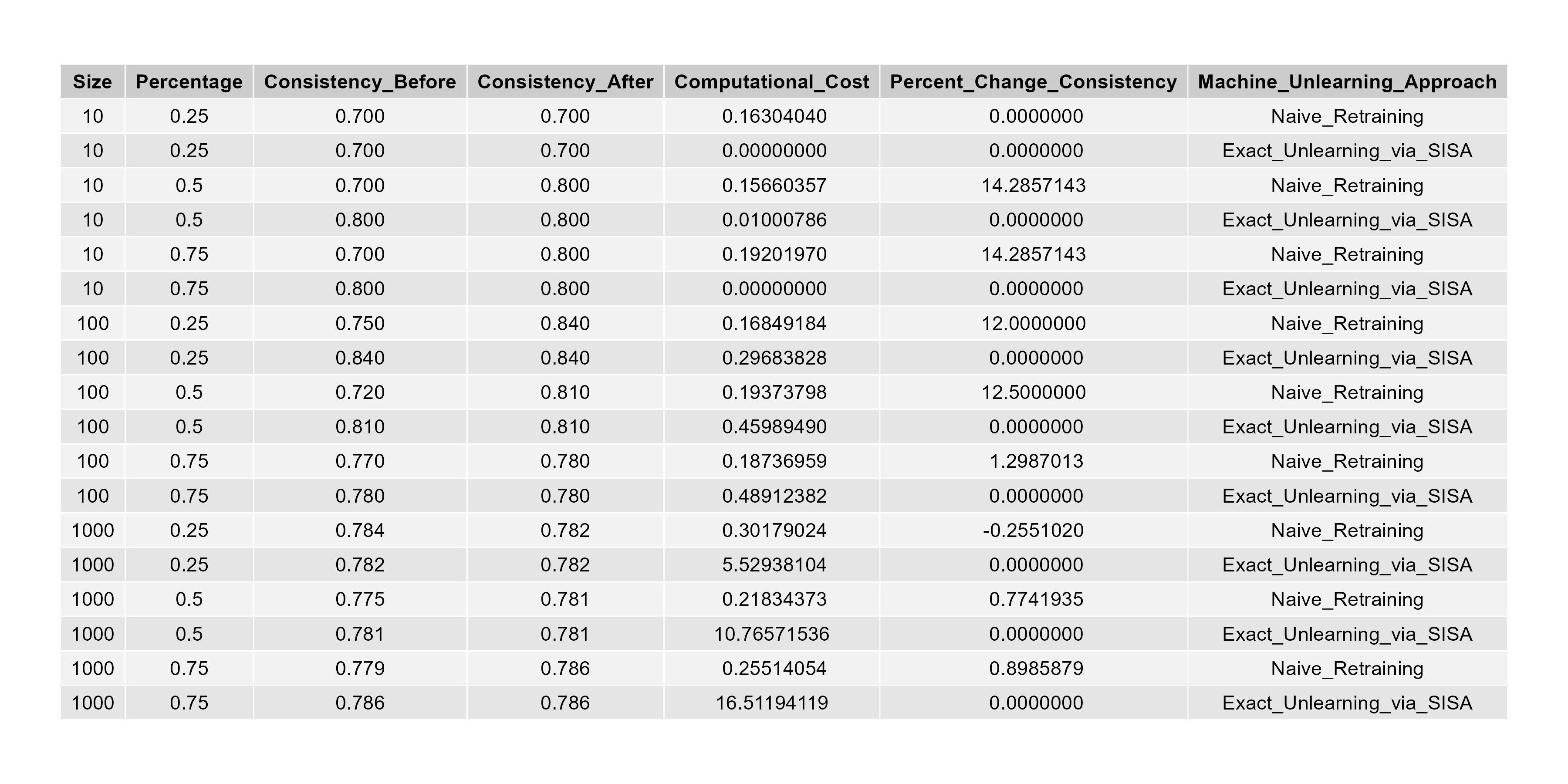}
                    \caption{Our Experimental Result Dataset in its Entirety}
                \end{figure}
        \end{enumerate}

\section{Results}
\label{sec:Results}

\begin{figure*}[h]
        \centering
        \includegraphics[width=1\linewidth]{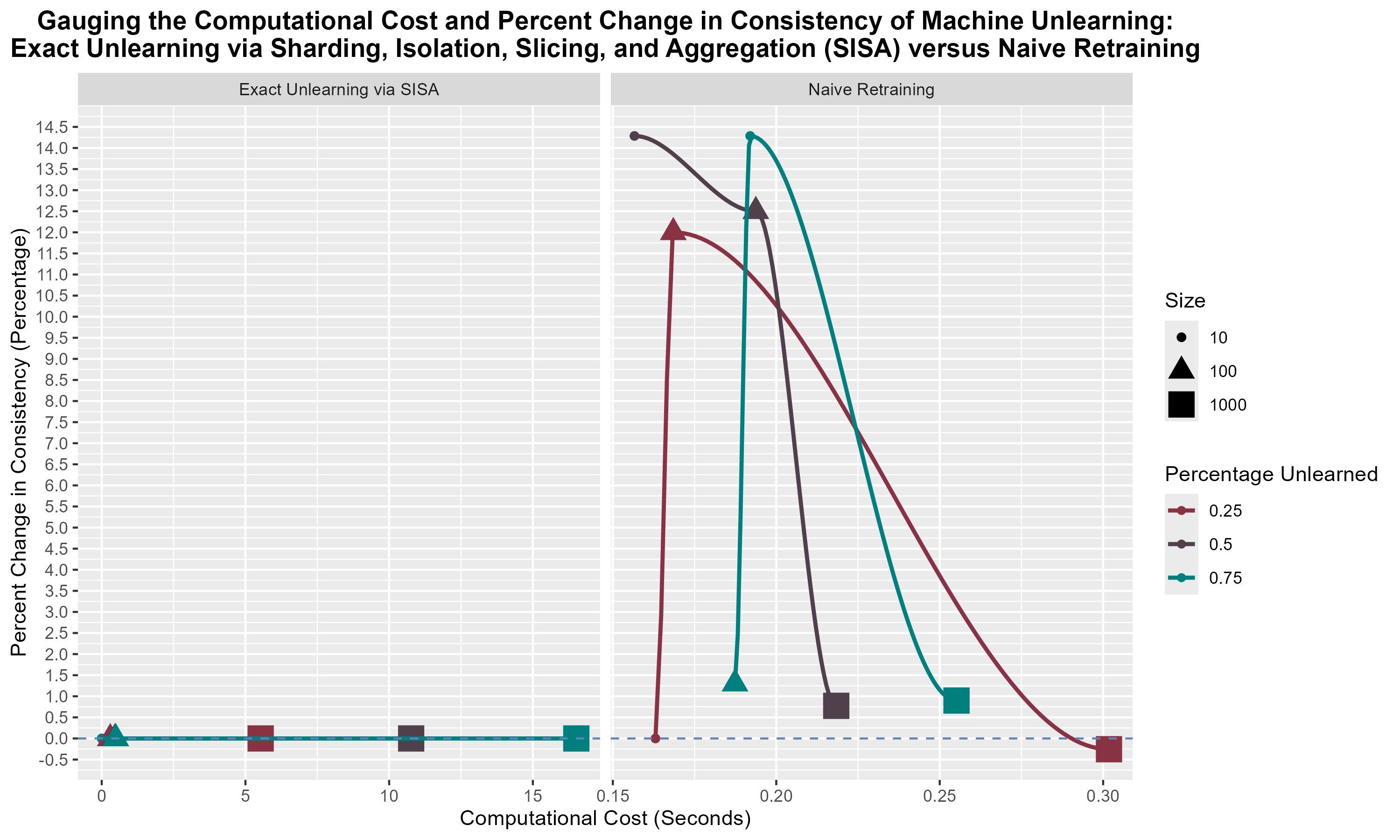}
        \caption{Gauging the Computational Cost and Percent Change in Consistency of Machine Unlearning}
        \label{fig:result_visualization}
    \end{figure*}

    As we move toward interpreting our results, we first provide what we believe is a succinct visual representation of our experimental outcomes (see Fig. \ref{fig:result_visualization}).
    
    Before we begin in earnest, we delineate a core concept of our graphical results.

    \begin{enumerate}[label=(Con. \arabic*), left=2em]
        \item \textit{A positive Percent Change in Consistency means that the model's predictive ability (or classifier accuracy) rose as a result of arbitrary deletions to its training set. Conversely, a negative Percent Change in Consistency would signify the degradation of a model's predictive ability.}
    \end{enumerate}
    
    With this in mind, the experimental outcome of Exact Unlearning via SISA aligns with our intuitive understanding of the Machine Unlearning framework. The specially configured Machine Learning model aims to permit element deletions without impacting the model's predictive ability. The Exact Unlearning via SISA facet of the provided graph is exceedingly self-explanatory. Seeing as how we implemented Exact Unlearning via SISA with a specific variant of Random Forest that facilitates data deletion (where the model can expunge elements used to train and fit the model), we can safely assume that the model's Consistency would remain constant. Indeed, the graph aligns with our assumption. For all instances of Exact Unlearning via SISA, the change in Consistency computed was zero, which means the Consistency values recorded before and after the unlearning remained unchanged. In spite of this, an obvious trend emerged for the Exact Unlearning via SISA examinations. As the percentage of the model's training set we deleted increased, so to did the Computational Cost. Furthermore, Computational Cost proportionally increases with the size of the training set examined.
    
    Having said that, the results of the Naive Retraining examinations are more nuanced. In terms of preliminary observations, the same relationship exists between Computational Cost and the size of the training set. An increase in size usually correlates with higher Computational Cost. There seems to be little correlation in terms of trends pertaining to the percentage of the training set deleted however.

    \begin{itemize}
        \item For \(0.25\) percent unlearned, the Percent Change in Consistency rises as the size increases from 10 to 100 but drastically diminishes when the size is increased from 100 to 1000. In fact, the recorded Percent Change in Consistency for size 1000 was negative. All other readings were positive or zero, meaning that Consistency usually increased once elements were deleted and the model was refitted and retrained.
        \item For \(0.50\) percent unlearned, the trend is more straightforward. As the training set size increased, the Percent Change in Consistency decreased but remained positive.
        \item For \(0.75\) percent unlearned, size 100 had the smallest Computational Cost followed by size 10 and then size 100. In this way, the arrangement of recorded Computational Cost for \(0.75\) percent unlearned in ascending order is: size 100, size 10, and size 1000. The trend within this sequence is thus. From size 100 to size 10, the Percent Change in Consistency sores and then plummets from size 10 to size 1000. Again, the Percent Change in Consistency remains positive.
    \end{itemize}
    
    All in all, we claim the following with confidence.

    \begin{enumerate}[label=(Cl. \arabic*), left=2em]
        \item \textit{Exact Unlearning via SISA, on average, demonstrated a higher Computational Cost when compared to Naive Retraining.}
    \end{enumerate}

    Therefore, our first hypothetical assertion (see Sec. \ref{subsec:Hypothesis}) was proven false. Based on the results, the process of unlearning for Exact Unlearning via SISA was slower than Naive Retraining. We suspect that the balancing feature of the specialized Random Forest model instance is the culprit. To maintain a consistent prediction ability, the model has to reconfigure and adjust a selection of its decision trees. In essence, there is more going on \say{under the hood} for the model tied to Exact Unlearning via SISA as opposed to Naive Retraining, which makes use of a standard Random Forest model.
    
    Likewise, our second hypothetical assertion was proven negative as well. 

    \begin{enumerate}[label=(Cl. \arabic*), left=2em, start=2]
        \item \textit{The consistency for both approaches was bounded within the range of \(0.7\) to \(0.84\), inclusive.}
    \end{enumerate}
    
    In fact, the Consistency readings before the unlearning process was enacted favored Exact Unlearning via SISA, for it produced a slightly more accurate prediction.
    
    At the end of the day, the choice of model and approach is apparent. As it relates to the ensemble variant of Machine Learning techniques, the better approach is Exact Unlearning via SISA, which employs a Removal-Enabled Random Forest or \textit{DaRE}. Granted there is a performance trade-off. To obtain consistent prediction ability, which our experiment demonstrated was the case for \textit{DaRE}, a commensurate depreciation in Computational Cost must be accepted. The process of unlearning will be a time-intensive process, especially as the size of the training set increases. Based on our findings, the only situation in which Naive Retraining would be appropriate is when dealing with sufficiently small datasets. As such, the Computational Cost will be negligible, but that is to be expected when dealing with small segments of data. From an asymptotic standpoint, we suspect that the perceived benefits of Naive Retraining would vanish as the size of the dataset under examination increases in an unbounded fashion.

\section{Review of Research and Future Work}
\label{sec:Review_of_Research}

    Incorporating a Machine Unlearning model such as \textit{DaRE}'s Random Forest into a PU Learning framework presents an intriguing possibility. While a concrete use case is not immediately apparent, the potential benefits of such a hybrid framework are evident. A classifier that operates on partially labeled data--data predominantly skewed toward labeled, positive samples--while also possessing the ability to unlearn erroneous or undesirable data points could address numerous challenges in real-world Machine Learning. These challenges include:
    
    \begin{enumerate}[label=(Chlg. \arabic*.),left=2em]
        \item \textit{Handling outdated examples,}
        \item \textit{Mitigating the impact of outliers,}
        \item \textit{Removing poisoned samples}, and
        \item \textit{Addressing noisy labels or harmful biases.}
    \end{enumerate}
    
    This study introduces the conceptual foundation for what we term the \textit{PUMU} framework: Positive Unlabeled Learning with Machine Unlearning. Such a framework could serve diverse applications, particularly in domains where data quality and ethical considerations are paramount, such as healthcare, fraud detection, and social media content moderation.
    
    Reflecting on the research conducted thus far, the potential for \textit{PUMU} as a novel framework is significant. PU Learning, known for its ability to classify unlabeled datasets that lack annotations, is particularly valuable in real-world scenarios where labeling data is impractical or expensive. The dual functionality of PU Learning and Machine Unlearning in a unified framework could address a wide range of challenges. For example, in fraud detection, data points flagged as fraudulent could be treated as positive instances, while Machine Unlearning ensures the removal of mislabeled or false positives dynamically. Similarly, in healthcare, \textit{PUMU} could enable robust diagnosis models that adaptively refine their training sets by forgetting outdated or erroneous medical records, preserving accuracy and ethical compliance. By extending PU Learning with Machine Unlearning, the resulting framework would not only enhance predictive performance in partially labeled datasets but also address the growing demand for data privacy and adaptability. Brainstorming additional scenarios where this framework could excel is an exciting avenue for future research.

\section{Conclusion}
\label{sec:Conclusion}

    Machine Unlearning can be likened to the \textit{Neuralyzer} from the movie \textit{Men in Black}. Just as the \textit{Neuralyzer} isolates and edits specific memories from its target, Machine Unlearning targets and erases the influence of specific data points within a Machine Learning model. This process ensures that the model effectively \say{forgets} the undesired data while retaining its overall knowledge and utility. Much like the \textit{Neuralyzer}’s ability to wipe memories seamlessly, Machine Unlearning aims to maintain the integrity of the model while addressing privacy and ethical concerns.
    
    Putting aside \say{pop culture} references, our study uncovered significant benefits of Machine Unlearning, particularly the superiority of Exact Unlearning via SISA when compared to Naive Retraining for large datasets. However, these benefits come with notable challenges. Below, we juxtapose the findings and contributions of this study with the challenges identified in prior research.
    
    \subsection{Study Findings}
    \label{subsec"Study_Findings}
    
        The study demonstrated that Exact Unlearning via SISA, as implemented in \textit{DaRE}, provides consistent prediction performance while enabling efficient unlearning. This approach is especially suitable for scenarios involving large datasets, where the computational cost of retraining entire models becomes prohibitive. By employing a shard-based strategy, SISA isolates the influence of individual data points, ensuring both accuracy and privacy compliance. On the other hand, Naive Retraining is only viable for small datasets due to its high computational cost for larger datasets. Our findings suggest that Naive Retraining’s perceived advantages diminish asymptotically as dataset size increases.
    
    \subsection{Challenges Identified}
    \label{subsec:Challenges_Identified}
    Despite its promise, Machine Unlearning faces several challenges:  
    
        \begin{enumerate}[label=(Chlg. \arabic*.),left=2em]
        \item \textbf{Data Dependencies}: \textit{Removing a single data point can disrupt the intricate relationships within the dataset, leading to unforeseen performance declines.}  
        \item \textbf{Model Complexity}: \textit{Large and complex models, such as deep neural networks, make it difficult to isolate and remove the influence of specific data points.} 
        \item \textbf{Computational Costs}: \textit{Iterative optimization processes involved in unlearning are time-intensive, particularly as models and datasets grow in size.}  
        \item \textbf{Privacy Risks}: \textit{During the unlearning process, the potential for information leakage through statistical anomalies or changes in model accuracy remains a concern.}  
        \item \textbf{Dynamic Datasets}: \textit{As datasets evolve over time, tracing and removing the influence of specific data points becomes increasingly complex.}  
        \item \textbf{Latency Concerns}: \textit{Unlearning processes may introduce delays that hinder prompt model updates required for low-latency predictions.}
        \end{enumerate}

These challenges, originally articulated by \textit{Jaman et al.} \cite{Jaman2024}, highlight the need for continued innovation in Machine Unlearning methods, particularly those that address the computational and privacy issues inherent in real-world applications. By advancing frameworks like \textit{PUMU} and refining unlearning algorithms, future research can help bridge the gap between theoretical promise and practical utility.

\bibliographystyle{IEEEtran}
\bibliography{references}

\begin{thebibliography}{10}
\providecommand{\url}[1]{#1}
\csname url@samestyle\endcsname
\providecommand{\newblock}{\relax}
\providecommand{\bibinfo}[2]{#2}
\providecommand{\BIBentrySTDinterwordspacing}{\spaceskip=0pt\relax}
\providecommand{\BIBentryALTinterwordstretchfactor}{4}
\providecommand{\BIBentryALTinterwordspacing}{\spaceskip=\fontdimen2\font plus
\BIBentryALTinterwordstretchfactor\fontdimen3\font minus \fontdimen4\font\relax}
\providecommand{\BIBforeignlanguage}[2]{{%
\expandafter\ifx\csname l@#1\endcsname\relax
\typeout{** WARNING: IEEEtran.bst: No hyphenation pattern has been}%
\typeout{** loaded for the language `#1'. Using the pattern for}%
\typeout{** the default language instead.}%
\else
\language=\csname l@#1\endcsname
\fi
#2}}
\providecommand{\BIBdecl}{\relax}
\BIBdecl

\bibitem{Mao2013}
H.~Mao, X.~Shuai, and A.~Kapadia, \emph{Loose Tweets: An Analysis of Privacy Leaks on Twitter}.\hskip 1em plus 0.5em minus 0.4em\relax ACM Digital Library, 2013.

\bibitem{Xu2024}
J.~Xu, Z.~Wu, C.~Wang, and X.~Jia, ``Machine unlearning: Solutions and challenges,'' \emph{IEEE Transactions on Emerging Topics in Computational Intelligence}, vol.~8, pp. 2150--2168, 6 2024.

\bibitem{Jaman2024}
L.~Jaman, R.~Alsharabi, and P.~M. Elkafrawy, ``Machine unlearning: An overview of the paradigm shift in the evolution of ai,'' in \emph{21st International Learning and Technology Conference: Reality and Science Fiction in Education, L and T 2024}.\hskip 1em plus 0.5em minus 0.4em\relax Institute of Electrical and Electronics Engineers Inc., 2024, pp. 25--29.

\bibitem{Xu2023}
H.~Xu, T.~Zhu, L.~Zhang, W.~Zhou, and P.~S. Yu, ``Machine unlearning: A survey,'' \emph{ACM Computing Surveys}, vol.~56, 1 2023.

\bibitem{Brophy2020}
\BIBentryALTinterwordspacing
J.~Brophy and D.~Lowd, ``Machine unlearning for random forests,'' 9 2020. [Online]. Available: \url{http://arxiv.org/abs/2009.05567}
\BIBentrySTDinterwordspacing

\bibitem{Sedhai2015}
S.~Sedhai and A.~Sun, ``Hspam14: A collection of 14 million tweets for hashtag-oriented spam research,'' in \emph{SIGIR 2015 - Proceedings of the 38th International ACM SIGIR Conference on Research and Development in Information Retrieval}.\hskip 1em plus 0.5em minus 0.4em\relax Association for Computing Machinery, Inc, 8 2015, pp. 223--232.

\bibitem{VoiceandView2024}
\BIBentryALTinterwordspacing
Voice, V.~Writers, and Staff, ``Infinity tweet secret stalkers: A dangerous online threat.'' [Online]. Available: \url{https://www.voiceandview.com/infinity-tweet-secret-stalkers-online-threat/}
\BIBentrySTDinterwordspacing

\bibitem{AHA2020}
\BIBentryALTinterwordspacing
A.~H.~A. Writers and Staff, ``Fbi psa: Child abductors potentially using social media or social networks to lure victims in lieu of an in-person ruse,'' 2020. [Online]. Available: \url{https://www.aha.org/other-cybersecurity-reports/2020-10-15-fbi-psa-child-abductors-potentially-using-social-media-or}
\BIBentrySTDinterwordspacing

\bibitem{Goggin2023}
\BIBentryALTinterwordspacing
B.~Goggin, ``A 13-year-old boy was groomed publicly on twitter and kidnapped, despite numerous chances to stop it,'' 2023. [Online]. Available: \url{https://www.nbcnews.com/tech/social-media/twitter-elon-musk-boy-kidnapped-groomed-discord-roblox-mcconney-rcna77985}
\BIBentrySTDinterwordspacing

\bibitem{Cao2015}
Y.~Cao and J.~Yang, ``Towards making systems forget with machine unlearning,'' in \emph{Proceedings - IEEE Symposium on Security and Privacy}, vol. 2015-July.\hskip 1em plus 0.5em minus 0.4em\relax Institute of Electrical and Electronics Engineers Inc., 7 2015, pp. 463--480.

\bibitem{WangkunXu2024}
W.~Xu and F.~Teng, ``Task-aware machine unlearning and its application in load forecasting,'' \emph{IEEE Transactions on Power Systems}, 2024.

\bibitem{Hu2024}
H.~Hu, S.~Wang, T.~Dong, and M.~Xue, ``Learn what you want to unlearn: Unlearning inversion attacks against machine unlearning,'' in \emph{2024 IEEE Symposium on Security and Privacy (SP)}.\hskip 1em plus 0.5em minus 0.4em\relax IEEE, 5 2024, pp. 3257--3275.

\bibitem{Zhang2024}
X.~Zhang, C.~Zhang, J.~Lou, K.~Wu, Z.~Wang, and X.~Chen, ``Duplexguard : Safeguarding deletion right in machine unlearning via duplex watermarking,'' \emph{IEEE Transactions on Dependable and Secure Computing}, pp. 1--15, 9 2024.

\bibitem{Lin2023}
H.~Lin, J.~W. Chung, Y.~Lao, and W.~Zhao, ``Machine unlearning in gradient boosting decision trees,'' in \emph{Proceedings of the ACM SIGKDD International Conference on Knowledge Discovery and Data Mining}.\hskip 1em plus 0.5em minus 0.4em\relax Association for Computing Machinery, 8 2023, pp. 1374--1383.

\bibitem{Tarun2023}
A.~K. Tarun, V.~S. Chundawat, M.~Mandal, and M.~Kankanhalli, ``Fast yet effective machine unlearning,'' \emph{IEEE Transactions on Neural Networks and Learning Systems}, 2023.

\bibitem{Schelter2021}
S.~Schelter, S.~Grafberger, and T.~Dunning, ``Hedgecut: Maintaining randomised trees for low-latency machine unlearning,'' in \emph{Proceedings of the ACM SIGMOD International Conference on Management of Data}.\hskip 1em plus 0.5em minus 0.4em\relax Association for Computing Machinery, 2021, pp. 1545--1557.

\bibitem{Juliussen2023}
B.~A. Juliussen, J.~P. Rui, and D.~Johansen, ``Algorithms that forget: Machine unlearning and the right to erasure,'' \emph{Computer Law and Security Review}, vol.~51, 11 2023.

\bibitem{Chen2025}
A.~Chen, Y.~Li, C.~Zhao, and M.~Huai, ``A survey of security and privacy issues of machine unlearning,'' \emph{AI Magazine}, vol.~46, 3 2025.

\bibitem{Brophy2021}
\BIBentryALTinterwordspacing
J.~Brophy, ``Dare rf: Data removal-enabled random forests,'' 2021. [Online]. Available: \url{https://github.com/jjbrophy47/dare_rf}
\BIBentrySTDinterwordspacing

\end{thebibliography}

\end{document}